\pdfoutput=1

\documentclass[11pt]{article}

\usepackage{coling}

\usepackage{times}
\usepackage{latexsym}
\usepackage{pifont}
\usepackage[T1]{fontenc}

\usepackage[utf8]{inputenc}
\usepackage{booktabs}
\usepackage{microtype}
\usepackage{xcolor}      
\usepackage{inconsolata}
\usepackage{multirow}
\usepackage{graphicx}
\usepackage{arydshln}
\usepackage{amsmath}
\usepackage{amssymb}
\usepackage{mathtools}
\usepackage{amsthm}

\definecolor{myyellow}{RGB}{235,184,57}
\definecolor{myred}{RGB}{240,73,110}
\definecolor{mygreen}{RGB}{12,215,159}
\title{Can Many-Shot In-Context Learning Help LLMs as Evaluators? \\ A Preliminary Empirical Study}

\author{Mingyang Song, Mao Zheng,  Xuan Luo, Yue Pan\\
	Tencent Hunyuan \\
	{\tt nickmysong@tencent.com} \\
}

\begin{document}
	\maketitle
	\begin{abstract}
Utilizing Large Language Models (LLMs) as evaluators to assess the performance of LLMs has garnered attention. However, this kind of evaluation approach is affected by potential biases within LLMs, raising concerns about the accuracy and reliability of the evaluation results of LLMs. To address this problem, we propose and study two many-shot In-Context Learning (ICL) prompt templates to help LLM evaluators mitigate potential biases: Many-Shot with Reference (MSwR) and Many-Shot without Reference (MSoR). Specifically, the former utilizes in-context examples with model-generated evaluation rationales as references, while the latter does not include these references. Using these prompt designs, we investigate the impact of increasing the number of in-context examples on the consistency and quality of the evaluation results. Experimental results show that advanced LLMs, such as GPT-4o, perform better in the many-shot regime than in the zero-shot and few-shot regimes. Furthermore, when using GPT-4o as an evaluator in the many-shot regime, adopting MSwR as the prompt template performs better than MSoR.
	\end{abstract}

	\section{Introduction}
	LLMs such as GPT-4o \cite{openai23}, Gemini1.5-Pro \cite{gemini}, and Claude3.5-Sonnet \cite{claude3} have demonstrated remarkable capabilities across a wide range of Natural Language Processing (NLP) tasks, becoming integral tools in various applications. The rapid advancement of LLMs \cite{palm} underscores the critical need to evaluate their alignment with human intent in generated responses. Therefore, evaluation has emerged as a crucial research area pivotal to the success of LLMs \cite{survey_eval}, especially for using LLMs as evaluators.
	
	LLMs like GPT-4 have shown exceptional performance across various tasks, leading to their wide adoption as both evaluators \cite{is_evaluator, GPTScore, wang2023, judge, unfair, chen} and annotators \cite{peng2023}. However, the robustness of LLMs as evaluators remains uncertain, given their sensitivity to textual instructions \cite{xu2023, Turpin23} and potential biases \cite{unfair, judge, chen}. To this end, researchers, such as \citet{unfair}, focus on addressing the potential biases that exist when LLMs act as evaluators.

	\begin{figure}
		\centering
		\includegraphics[scale=0.34]{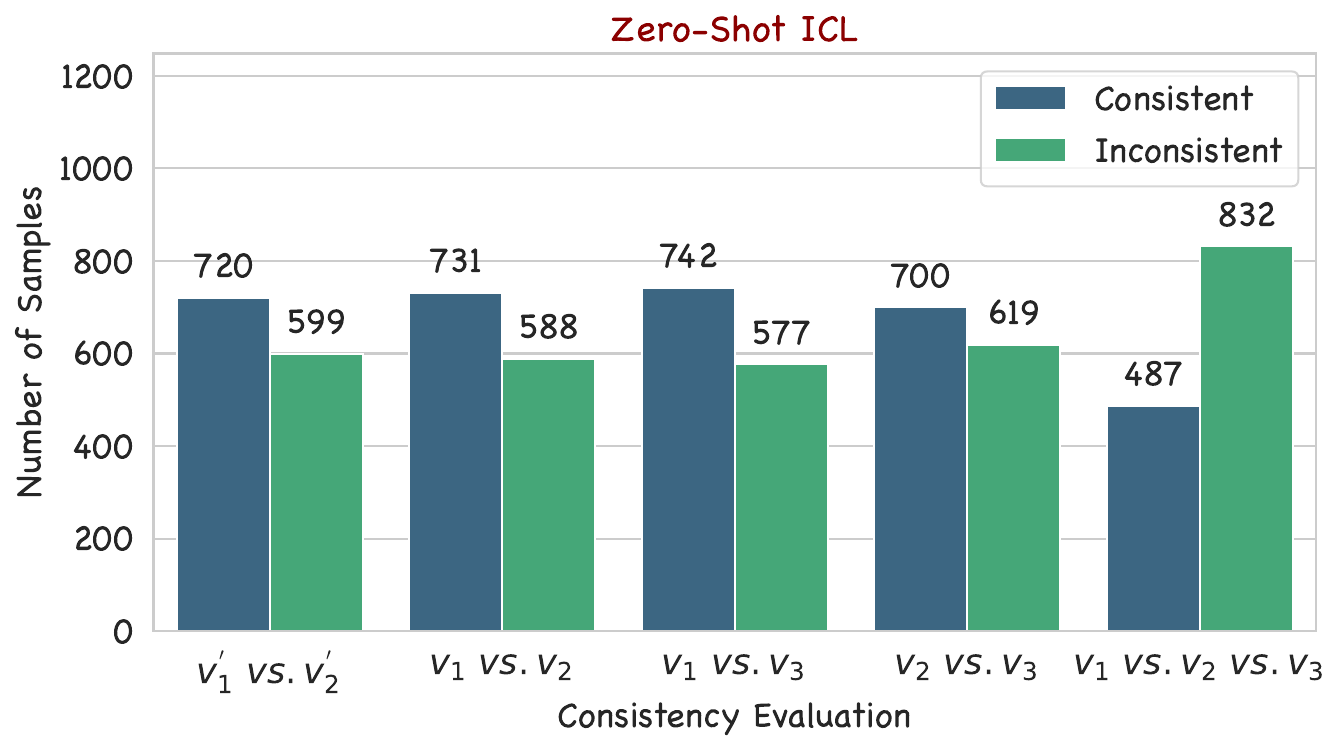}
		\caption{Consistency between different versions of evaluation results by adopting GPT-4o as a zero-shot evaluator. $v^{\prime}_1$ and $v^{\prime}_2$ are the results based on Prompt(A) in Table~\ref{zero_shot}. $v_1$, $v_2$, and $v_3$ are results based on Prompt(B) in Table~\ref{zero_shot}. Prompts (A) and (B) differ in whether to output the rating first or later. The consistency evaluations show that Prompt (A) and (B) almost obtain the agreement results, but the latter is convenient for constructing many-shot in-context examples, so we adopt the latter generated rationales in this study. $v_1$ vs. $v_2$ denotes comparing the first and second versions of evaluations. $v_1$ vs. $v_2$ vs. $v_3$ denotes the comparison between the three versions of evaluations.}
		\label{intro}
	\end{figure}
	
	\begin{table*}[h]
		\centering
		\tiny
		
		\renewcommand\arraystretch{1.15}
			\resizebox{0.99\linewidth}{!}{

				\begin{tabular}[t]{@{}p{0.8\linewidth}@{}}
					\toprule

					\textbf{Prompt(A).} \textit{The zero-shot prompt for single answer grading in \citet{judge}.} \\\hline
					
					\textit{Please act as an impartial judge and evaluate the quality of the response provided by an AI assistant to the user question displayed below. Your evaluation should consider factors such as the helpfulness, relevance, accuracy, depth, creativity, and level of detail of the response. Begin your evaluation by providing a short explanation. Be as objective as possible. After providing your explanation, please rate the response on a scale of 1 to 10 by strictly following this format: "[[rating]]", for example: "Rating: [[5]]".}\\
					\textbf{Question}\\
					\{question\}\\
					\textbf{Response}\\
					\{response\}\\
					\midrule
					
					\begin{tabular}[t]{@{}p{\linewidth}@{}}
						
						\textbf{Prompt(B).} \textit{The zero-shot prompt for single answer grading in this paper.} \\\hline
						
						\textit{Please act as an impartial judge and evaluate the quality of the response provided by an AI assistant to the user question displayed below. Your evaluation should consider factors such as the helpfulness, relevance, accuracy, depth, creativity, and level of detail of the response. Please rate the response on a scale of 1 to 10 by strictly following this JSON format: \{"rating":"", "reason":""\}. The "rating" should be as objective as possible. The "reason" denotes a comprehensive explanation of your rating, which should consider factors such as helpfulness, relevance, accuracy, depth, creativity, and level of detail of the response should also be considered. Only return the JSON results and do not give any explanation.}\\
						\textbf{Question}\\
						\{question\}\\
						\textbf{Response}\\
						\{response\}\\
						
					\end{tabular}
					\\
					\bottomrule
					
			\end{tabular}}
			\caption{Two versions of zero-shot prompts are used for evaluation in this paper.}
			\label{zero_shot}
		\end{table*}
		
		Newly expanded context windows of LLMs allow researchers to investigate ICL with more shots than the zero-shot and few-shot regimes, namely many-shot ICL. To fully investigate the many-shot ICL, \citet{many_shot} explore the impact of the number of in-context examples by scaling shots to hundreds or thousands and find that many-shot can better reduce the biases of LLMs than the few-shot regime. Specifically, they show that using the many-shot in-context examples with chain-of-thought rationales generated through the zero-shot regime is effective, and the many-shot ICL may overcome the biases of LLMs, whereas few-shot ICL struggles.
		Therefore, the intuitive idea is to use the many-shot ICL, allowing LLMs as evaluators to see the zero-shot evaluations of similar questions and answers first and then scoring examples before scoring. 
		Therefore, an interesting question arises:
		\begin{itemize}
			\item \textit{Can many-shot in-context learning help LLMs as evaluators?}
		\end{itemize}
		
		Motivated by prior findings and the above issue, we preliminary verify the consistency of the widely used prompts of using LLMs as evaluators \cite{judge}, as shown in Table~\ref{zero_shot}. Concretely, the consistency experiments are based on the entire test set of GSM8K \cite{gsm8k}, where the inference answers are obtained based on LLaMA3-70B (the details are presented in $\S$~\ref{exp}). Figure~\ref{intro} presents that the consistency between the two versions of evaluations is low, with nearly half of the ratings being inconsistent.
		
		Inspired by the above question, in this paper, we investigate whether many-shot in-context learning helps LLMs as evaluators.
		Specifically, we introduce two versions of prompts for LLM evaluators, \textbf{M}any-\textbf{S}hot \textbf{w}ith \textbf{R}eference (\textbf{MSwR}) and \textbf{M}any-\textbf{S}hot with\textbf{o}ut \textbf{R}eference (\textbf{MSoR}). The former utilizes in-context examples with model-generated rationales as guidance, and the latter removes rationales used in the former. Meanwhile, we reveal the symbol bias in LLMs and explore a simple approach for mitigating this issue.
		Experiments show that many-shot ICL can help GPT-4o-as-a-Jduge obtain higher quality and consistent evaluation results. As the number of in-context examples increases, the quality and consistency of evaluation improves significantly. Furthermore, we further verify the effectiveness of the proposed simple yet effective approach for mitigating the symbol bias in pairwise comparison of GPT-4o as an evaluator. To the best of our knowledge, we are the first to attempt to study LLM as an evaluator using the many-shot ICL regime.
		
		\begin{figure*}[t]
			\centering
			\includegraphics[scale=0.31]{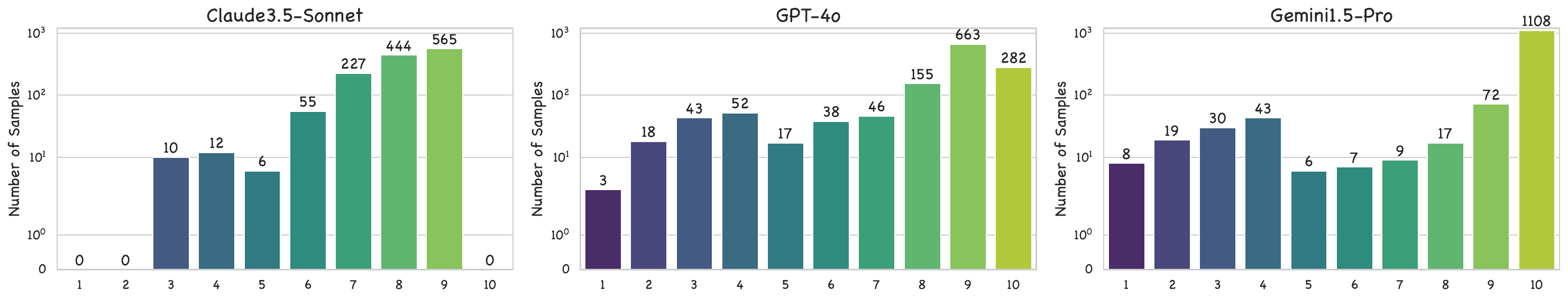}
			\caption{Evaluate the results of LLaMA3-70b on the GSM8K dataset using the Prompt(A).}
			\label{answer_dis}
		\end{figure*}
		\begin{figure*}[t]
			\centering
			\includegraphics[scale=0.31]{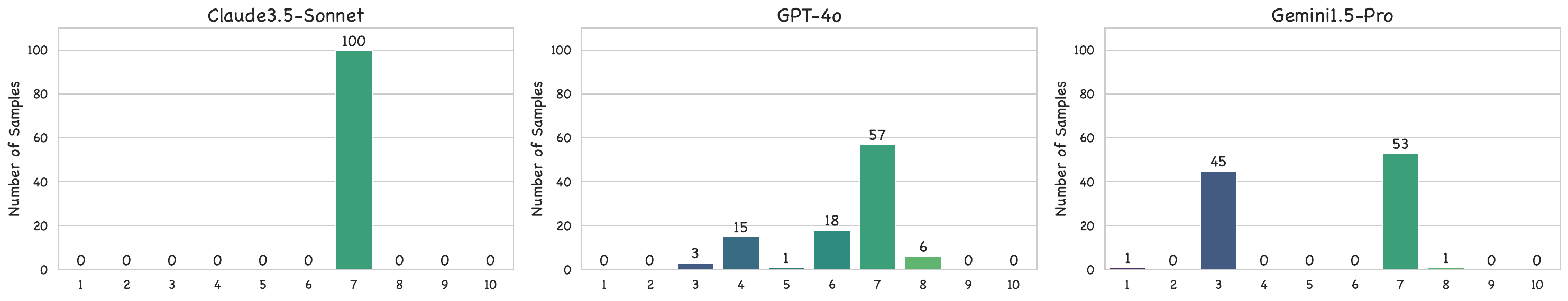}
			\caption{Results of random selection.}
			\label{answer_dis_random}
		\end{figure*}

		\section{Methodology}
		\subsection{Background of Many-Shot ICL}
		LLMs excel at few-shot in-context learning, which involves learning from a few input-output demonstrations (“shots”) provided in context at inference without weight updates \cite{few_shot}. Newly expanded long-context LLMs allow us to investigate ICL with hundreds of in-context examples \cite{many_demo, many_shot}. 
		\subsection{Recalling Potential Biases in LLMs}
		LLMs as evaluators possess potential biases, which have been widely explored by \citet{unfair, wu, judge, chen, selection_bias}. For example, positional bias in LLMs refers to the phenomenon where, during pairwise comparisons, LLM evaluators tend to favor one side of a pair regardless of the actual quality of the answers. In this paper, we investigate whether leveraging many-shot ICL helps LLMs as better evaluators. 
		
		To select the tested LLM and reveal the selection bias, we conduct a simple symbol selection experiment (set temperature=0.7 for all tested LLMs), using the designed prompt in Table~\ref{random_prompt} to let LLMs randomly choose a symbol. The test results are shown in Figure~\ref{answer_dis_random}, and from these results, it can be observed that different LLMs tend different symbols.
		Meanwhile, we also evaluate the performance of LLaMA3-70B on the GSM8K dataset using Prompt(A) in Table~\ref{zero_shot}  as shown in Figure~\ref{answer_dis}. From the results in Figure~\ref{answer_dis_random} and Figure~\ref{answer_dis}, we observe that the results of GPT-4o are more reasonable than the others. Moreover, previous studies have demonstrated that GPT4 series models have the highest consistency with human evaluation.
		\subsection{Two Prompt Templates}
		In this section, we introduce two prompt templates for employing LLM as an evaluator: Many-Shot with Reference (MSwR) and Many-Shot without Reference (MSoR).
		\begin{table}[t]
			\centering
			\scriptsize
			
			\renewcommand\arraystretch{1.1}
				\resizebox{0.99\linewidth}{!}{

					\begin{tabular}[t]{@{}p{0.8\linewidth}@{}}
						\toprule
						\textit{Do not ask why, randomly select one from the following symbols. Just return the symbol you choose, no explanation is needed.} \\
						\\
						1 2 3 4 5 6 7 8 9 10\\
						\bottomrule
						
				\end{tabular}}
				\caption{Prompt template for testing selection bias.}
				\label{random_prompt}
			\end{table}
			\subsubsection{Many-Shot with Reference}
			Usually, when asking LLMs to select a score from 1-10 for evaluating answers to questions, providing reference answers may improve the evaluation quality of LLMs \cite{unfair, judge}.
			In the zero-shot ICL regime, no in-context examples are provided for the GPT-4o evaluator, which selects scores only depending on itself. In the few-shot ICL regime, a few in-context examples are provided for the GPT-4o evaluator. In the many-shot ICL regime, many in-context examples are provided for the GPT-4o evaluator. The differences among the three regimes above are as follows: (1) zero-shot, which lacks any reference information, causes LLM to score entirely based on its preferences. (2) few-shot, due to providing only a small number of examples, may lead to evaluation results that lack diversity. (3) many-shot, by offering a larger number of examples, can ensure diversity and quality in evaluations.
			Therefore, we present the prompt template MSwR, which uses model-generated rationales as the in-context examples. Specifically, MSwR-4 indicates the prompt template of MSwR using 4 shots with reference as a demonstration, as shown in Table~\ref{reinforced_icl}.
			\subsubsection{Many-Shot without Reference}
			Previous studies \cite{many_shot} find that in-context examples may limit the problem-solving approach of LLMs, so we propose a prompt template that does not need a reference for each in-context example. The designed prompt contains three parts: (1) a preamble, such as, “You will be provided questions similar to the ones below:”; (2) a list of unsolved inputs or problems.; (3) a few-shot prompt with outputs for the desired output format. Specifically, MSoR-(K=4)-4 indicates that using 4 shots without reference and 4 shots with reference, as shown in Table~\ref{unsupervised}.

			\begin{table}[t]
				\centering
				\tiny
				
				\renewcommand\arraystretch{1.1}
					\resizebox{0.99\linewidth}{!}{
						\begin{tabular}[t]{@{}p{0.8\linewidth}@{}}
							\toprule
							\begin{tabular}[t]{@{}p{\linewidth}@{}}
								\textit{Please act as an impartial judge and evaluate the quality of the response provided by an AI assistant to the user question displayed below. Your evaluation should consider factors such as the helpfulness, relevance, accuracy, depth, creativity, and level of detail of the response. Please rate the response on a scale of 1 to 10 by strictly following this JSON format: \{"rating":"", "reason":""\}. The "rating" should be as objective as possible. The "reason" denotes a comprehensive explanation of your rating, which should consider factors such as helpfulness, relevance, accuracy, depth, creativity, and level of detail of the response should also be considered. Only return the JSON results and do not give any explanation.}
								
								\textit{Now, I am going to give you a series of demonstrations of <Question>, <Response>, and Evaluations. When you respond, respond only with the Evaluation of the final pair of <Question> and <Response>, thinking step by step.}
								
								---\\
								{<Question>}\\
								\{question1\}\\
								{<Response>}\\
								\{response1\}\\
								\textbf{Evaluation}\\
								\{evaluation1\}\\
								---\\
								...\\
								---\\
								{<Question>}\\
								\{question4\}\\
								{<Response>}\\
								\{response4\}\\
								\textbf{Evaluation}\\
								\{evaluation4\}\\
								---\\
								{<Question>}\\
								\{final\_question\}\\
								{<Response>}\\
								\{final\_response\}\\
								\textbf{Evaluation}
							\end{tabular}
							\\
							\bottomrule
					\end{tabular}}
					\caption{Example of the prompt MSwR-(K=4).}
					\label{reinforced_icl}
				\end{table}
				\begin{table}[h!]
					\centering
					\tiny
					
					\renewcommand\arraystretch{1.1}
						\resizebox{0.99\linewidth}{!}{
							\begin{tabular}[t]{@{}p{0.8\linewidth}@{}}
								\toprule
								
								\begin{tabular}[t]{@{}p{\linewidth}@{}}
									\textit{Please act as an impartial judge and evaluate the quality of the response provided by an AI assistant to the user question displayed below. Your evaluation should consider factors such as the helpfulness, relevance, accuracy, depth, creativity, and level of detail of the response. Please rate the response on a scale of 1 to 10 by strictly following this JSON format: \{"rating":"", "reason":""\}. The "rating" should be as objective as possible. The "reason" denotes a comprehensive explanation of your rating, which should consider factors such as helpfulness, relevance, accuracy, depth, creativity, and level of detail of the response should also be considered. Only return the JSON results and do not give any explanation.}\\
									\textit{Now, I am going to give you a series of demonstrations of <Question> and <Response>.}\\
									---\\
									{<Question>}\\
									\{question1\}\\
									{<Response>}\\
									\{response1\}\\
									---\\
									...\\
									---\\
									{<Question>}\\
									\{question4\}\\
									{<Response>}\\
									\{response4\}\\
									---\\
									\textit{Now, I am going to give you a series of demonstrations of <Question>, <Response>, and Evaluations. When you respond, respond only with the Evaluation of the final pair of <Question> and <Response>, thinking step by step.}\\
									---\\
									{<Question>}\\
									\{question1\}\\
									{<Response>}\\
									\{response1\}\\
									\textbf{Evaluation}\\
									\{evaluation1\}\\
									---\\
									...\\
									---\\
									{<Question>}\\
									\{question4\}\\
									{<Response>}\\
									\{response4\}\\
									\textbf{Evaluation}\\
									\{evaluation4\}\\
									---\\
									{<Question>}\\
									\{final\_question\}\\
									{<Response>}\\
									\{final\_response\}\\
									\textbf{Evaluation}
								\end{tabular}
								\\
								\bottomrule
						\end{tabular}}
						\caption{Example of the prompt MSoR-(K=4)-4.}
						\label{unsupervised}
					\end{table}
					\begin{figure*}
						\centering
						\includegraphics[scale=0.31]{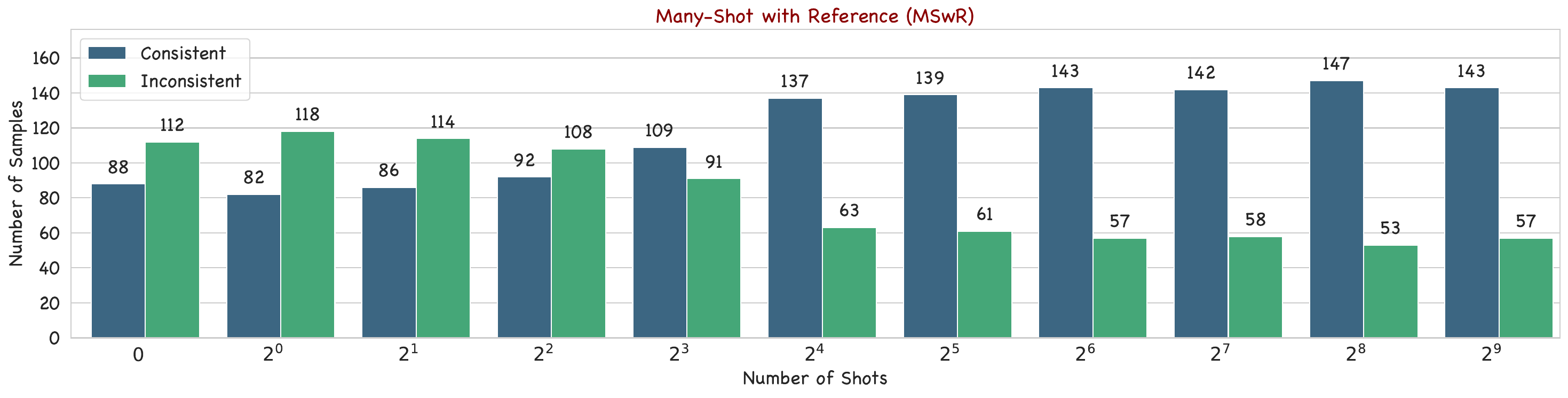}
						\caption{Consistency between two versions of evaluation results. Concretely, the bar corresponding to "0" on the x-axis represents the number of samples with consistent and inconsistent ratings in comparing evaluation results obtained twice using GPT-4o as the evaluator in the zero-shot regime. In addition, the zero-shot generated rationales are used for MSwR and MSoR. The bar corresponding to "$2^n$" on the x-axis represents the consistency of using the GPT-4o as an evaluator in MSwR.}
						\label{comp_long_gpt4}
					\end{figure*}
					\begin{figure*}[t!]
						\centering
						\includegraphics[scale=0.25]{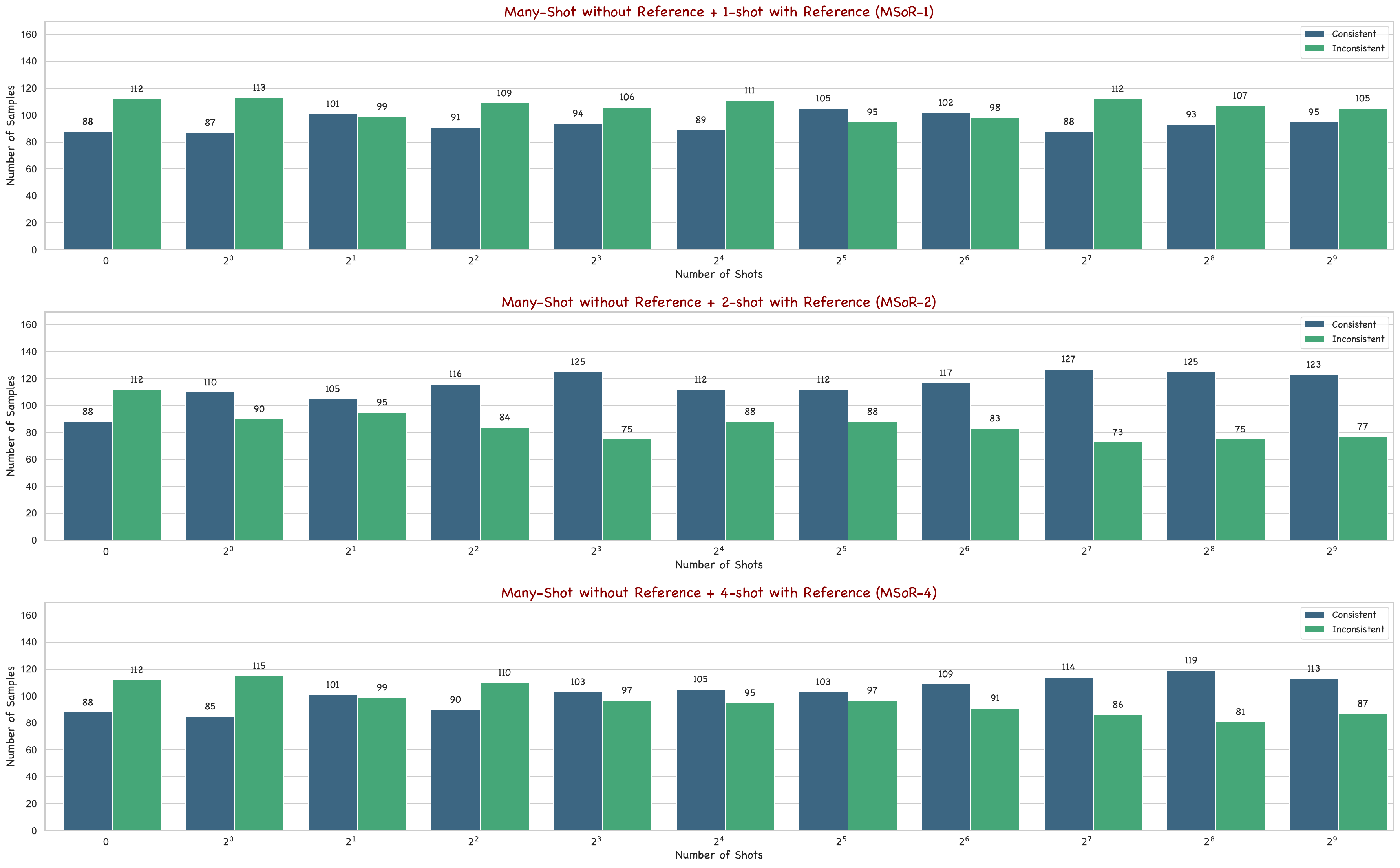}
						\caption{Compare the consistency of the results from the two evaluations.}
						\label{comp_unsupervised_shot}
					\end{figure*}
					\section{Experiments}
					\label{exp}
					\subsection{Experimental Settings}
					In our experiments, we use LLaMA3-70B to generate answers for each question in GSM8K \cite{gsm8k} with a temperature of 0.7. GSM8K\footnote{\url{https://huggingface.co/datasets/openai/gsm8k}} is previously introduced by \citet{gsm8k}, which comprises 8.5K high-quality grade school math problems meticulously crafted by human problem writers. This dataset is divided into 7.5K training and 1K test problems. Specifically, \textbf{each problem typically requires between 2 to 8 steps to solve}, primarily involving a sequence of elementary calculations using basic arithmetic operations (addition, subtraction, multiplication, and division). The problems are designed so that a proficient middle school student can solve each one. Furthermore, the problem-solving task requires models to solve problems with model-generated rationales, which may be challenging to evaluate. Then, we use LLaMA3-70B\footnote{\url{https://github.com/meta-llama/llama3}} to create model-generated rationales for GSM8K.
					There are two reasons for using LLaMA3 to infer the answers to the problems in GSM8K. First, the quality of answers obtained by LLaMA3 may be high or low, which makes it beneficial to use these answers to analyze the GPT-4o as an evaluator. Second, if GPT-4o is adopted to infer the answers, using GPT-4o to evaluate again has potential biases.
					We use the training set of GSM8K as the sampling pool of in-context samples and use the first 200 samples of the test set in GSM8K for experiments in this paper. Specifically, we randomly sample in-context examples for each ${K}$-shot prompt in each test data for reliable results. To ensure that using more shots provides additional information, any ${K}$-shot ICL prompt in our experiments includes all in-context examples from prompts with less than ${K}$ examples (all examples in the training set).

					\begin{table*}[t!]
						\scriptsize
						\centering
						\renewcommand\tabcolsep{11.7pt}
						\renewcommand\arraystretch{2.3}
						\begin{tabular}{ccccccccccc}
							\hline\hline
							
							\multicolumn{1}{c}{}     & $2^0$ & $2^1$ & $2^2$ & $2^3$ & $2^4$ & $2^5$ & $2^6$ & $2^7$ & $2^8$ & $2^9$\\\hline
							\multicolumn{11}{l}{Average Context Length in \textbf{Character} of A Single Test Sample} \\\hline
							MSwR-K  & 0.7K  &  1.1K   &  2.2K   &  3.6K   &   6.5K  &   13.6K  &  26.3K   &  50.7K &  98.3K &  198.5K\\
							MSoR-K-4 &  2.2K &   2.6K   & 3.1K &  4.6K   &  6.9K &  11.7K  & 20.9K   & 40.8K  & 81.3K & 159.7K\\
							\hline\hline   
						\end{tabular}
						\caption{Context length in tokens for K-shot MSwR-K and MSoR-K-4 (K = $2^0$, $2^1$, $2^2$, $2^3$, $2^4$, $2^5$, $2^6$, $2^7$, $2^8$, $2^9$).}
						\label{length}
					\end{table*}
					\begin{table*}[h!]
						\centering
						\tiny
						\renewcommand\arraystretch{1.4}
						\resizebox{0.99\linewidth}{!}{
							\begin{tabular}[t]{@{}p{0.8\linewidth}@{}}
								\toprule
								\textit{Please act as an impartial judge and evaluate the quality of the responses provided by two AI assistants to the user question displayed below. You should choose the assistant that follows the user’s instructions and answers the user’s question better. Your evaluation should consider factors such as the helpfulness, relevance, accuracy, depth, creativity, and level of detail of their responses. Begin your evaluation by comparing the two responses and provide a short explanation. Avoid any position biases and ensure that the order in which the responses were presented does not influence your decision. Do not allow the length of the responses to influence your evaluation. Do not favor certain names of the assistants. Be as objective as possible. After providing your explanation, output your final verdict by strictly following this format: "[[A]]" if assistant A is better, "[[B]]" if assistant B is better, and "[[C]]" for a tie.}\\
								\textbf{Problem}\\
								\textit{Please act as an impartial judge and evaluate the quality of the response provided by an AI assistant to the user question displayed below. Your evaluation should consider factors such as the helpfulness, relevance, accuracy, depth, creativity, and level of detail of the response. Please rate the response on a scale of 1 to 10 by strictly following this JSON format: \{"rating":"", "reason":""\}. The "rating" should be as objective as possible. The "reason" denotes a comprehensive explanation of your rating, which should consider factors such as helpfulness, relevance, accuracy, depth, creativity, and level of detail of the response should also be considered. Only return the JSON results and do not give any explanation.}\\
								\textbf{Question}\\
								\{question\}\\
								\textbf{Response}\\
								\{response\}\\
								\hline
							\end{tabular}
						}
						\resizebox{0.99\linewidth}{!}{
							\begin{tabular}[t]{@{}p{0.2\linewidth}p{0.2\linewidth}p{0.2\linewidth}p{0.2\linewidth}@{}}
								
								\textbf{The Assistant A’s Answer}& \textbf{The Assistant A’s Answer}& \textbf{The Assistant {\color{red}B}’s Answer}&\textbf{The Assistant {\color{red}B}’s Answer}\\
								
								\{Answer-A\}& \{Answer-B\}&\{Answer-A\}&\{Answer-B\}\\
								
								\textbf{The Assistant B’s Answer}& \textbf{The Assistant B’s Answer}& \textbf{The Assistant {\color{red}A}’s Answer}&\textbf{The Assistant {\color{red}A}’s Answer}\\
								
								\{Answer-B\}& \{Answer-A\}&\{Answer-B\}&\{Answer-A\}\\\hline
								
								\textit{Compare(A, B)} & \textit{Compare(B, A)} & \textit{Compare(A$^\dagger$, B$^\dagger$)}  & \textit{Compare(B$^\dagger$, A$^\dagger$)}\\
								\bottomrule
							\end{tabular}
						}
						\caption{The prompts used for pairwise comparison.}
						\label{compare_prompt}
					\end{table*}
					
					Inspired by Counting-Stars \cite{song2024countingstars}, even if too many shots are provided, the GPT-4o may not be able to utilize all of them. Because, in the Counting-Stars benchmark, when the number of pieces of evidence reaches 32, LLMs (e.g., GPT-4 Turbo, Gemini 1.5 Pro, and Claude3 Opus) may no longer accurately obtain all of them. Therefore, adding more many-shot in-context examples is probably not captured by LLMs for learning as a reference to evaluate. However, the many-shot regime and Counting-Stars are substantially different, so there is no noise from "haystack". Hence, we set the maximum number of the in-context examples to 128, i.e., 128-shots.
					
					In the experiments, we use GPT-4o with public API access, and the specific endpoint is “gpt-4o-2024-05-13”. We use Claude3.5-Sonnet with public API access, and the specific endpoint is “claude-3-5-sonnet-20240620”. For the convenience of introduction, when comparing zero-shot and many-shot regimes, we uniformly refer to the few-shot and many-shot regimes as the many-shot regime. The context length in character-level of a single test sample with $K$-shot in-context examples is presented in Table~\ref{length}.
					

					\subsection{Consistency Evaluation}
					We investigate the consistency between different versions of evaluation results generated by the GPT-4o evaluator. Here, the single answer grading evaluation results can be used to compare the consistency between different versions. As shown in Figure~\ref{comp_long_gpt4}, the bar corresponding to "$2^n$" on the x-axis represents the number of samples with consistent and inconsistent ratings in comparing evaluation results obtained twice using GPT-4o as an evaluator in MSwR. From the results, consistency improves as we increase the number of shots provided as in-context examples during inference.
					
					\begin{figure*}[t!]
						\centering
						\includegraphics[scale=0.32]{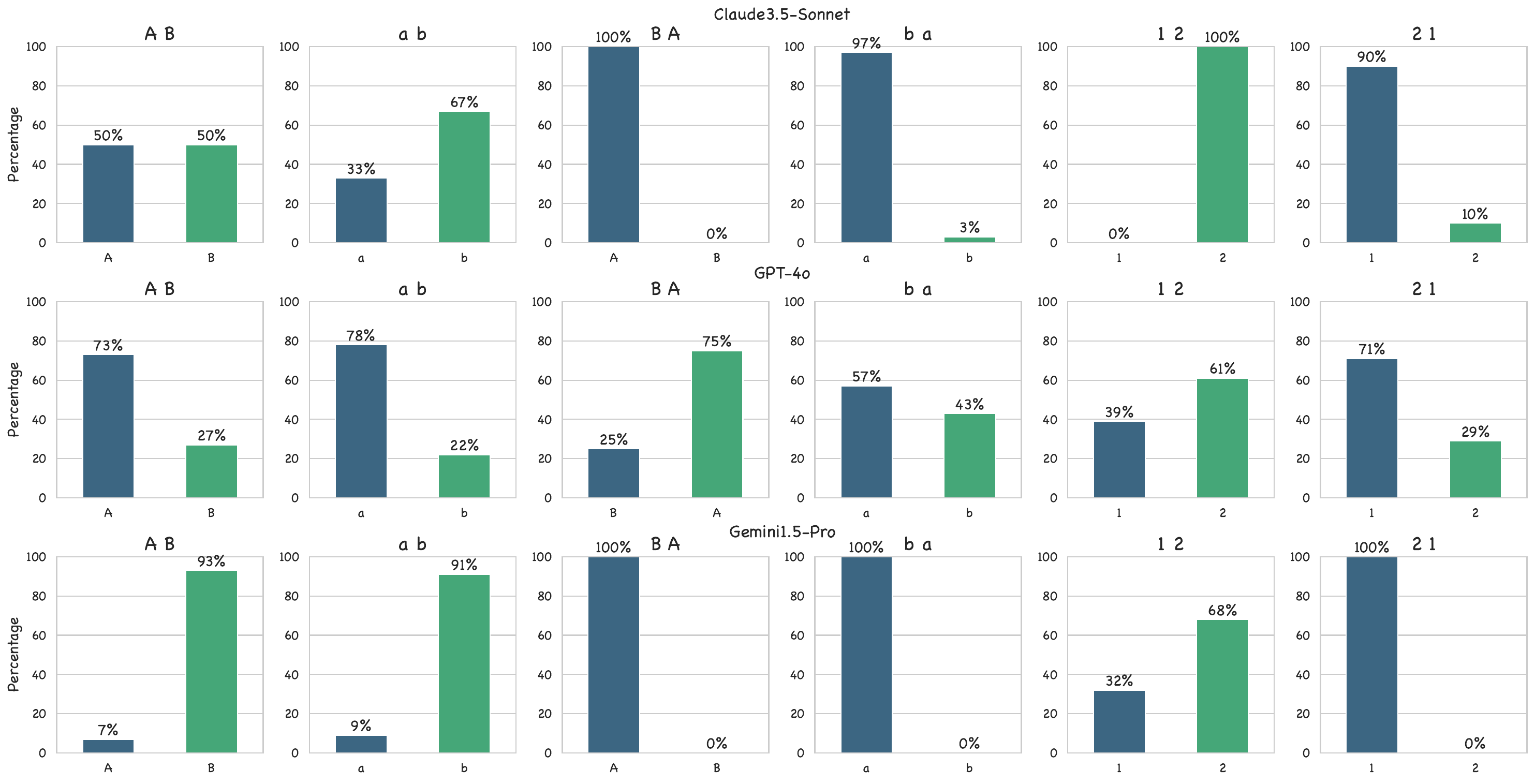}
						\caption{Results of randomly selecting different symbols.}
						\label{random_pair}
					\end{figure*}
					Recent studies \cite{unfair, judge, chen} have shown that the performance of GPT-4 as an evaluator is highly in agreement with those of humans. However, both human and LLM evaluators are subject to potential biases \cite{chen}.
					By analyzing prior studies, we suppose it unnecessary to obtain utterly accurate evaluation results (because this is difficult) by using LLMs as evaluators. It is only required to ensure that the evaluation results are highly consistent multiple times so that the single answer grading of GPT-4o as an evaluator may be effective. From all results in this work, we find that the many-shot ICL examples help the evaluation of LLMs more consistently, which is essential. We consider that the main reason may be that the many-shot in-context examples mitigate the potential biases of the GPT-4o evaluator.
					
					Meanwhile, we also implemented experiments via the prompt template MSoR, but the results show that this regime may be unsuitable in the scenario of acting as an evaluator because the problem of each sample is the same, but the scored questions and answers are different. From the results in Figure~\ref{comp_unsupervised_shot}, it can be seen that the consistency corresponds to the number of appended in-context samples with model-generated rationales but nearly not to the number of in-context examples without model-generated rationales.
					
					In addition, both the few-shot and many-shot regimes are sensitive to the selection and order of in-context examples. In the experimental setting of this paper, almost no test data will have the same in-context examples. Even under this condition, the evaluation results also show a high consistency, demonstrating the effectiveness of many-shot ICL in helping GPT-4o evaluator.

					\begin{figure*}[h!]
						\centering
						\includegraphics[scale=0.325]{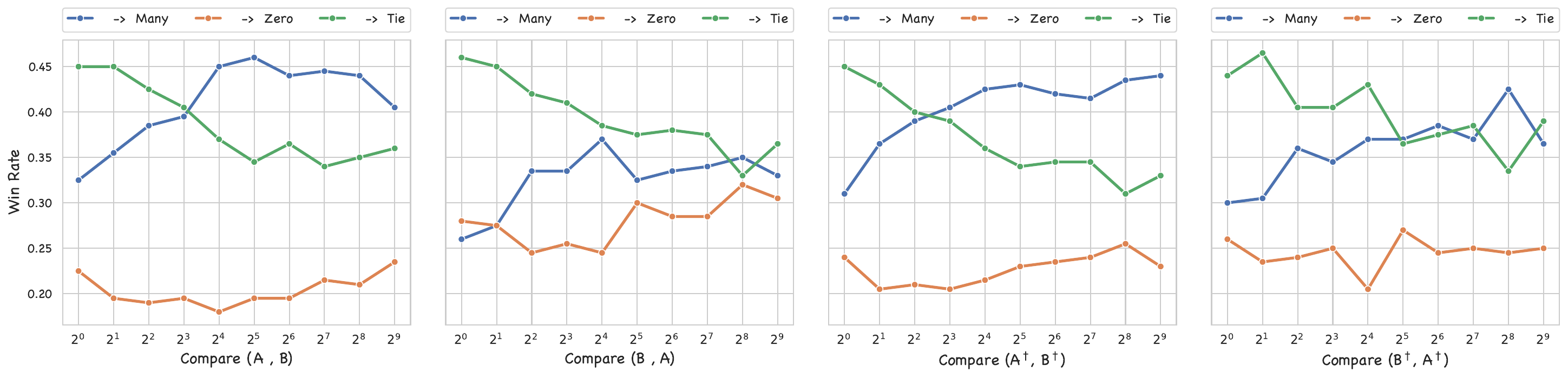}
						\caption{Win rate compared between zero-shot and many-shot regimes.}
						\label{comp_all}
					\end{figure*}
					
					\subsection{Quality Evaluation}
					After the consistency evaluation, an important question arises about: 
					\begin{itemize}
						\item \textit{{Does a high consistency refer to high-quality evaluations?}} 
					\end{itemize}
					
					Around this issue, we pairwise compare model-generated evaluation rationales between zero-shot and many-shot regimes using the designed prompts shown in Table~\ref{compare_prompt}.  
					In pairwise comparison, the GPT-4o evaluator is presented with a question and two answers and tasked with \textit{determining which is better or declaring a tie}.
					From Figure~\ref{comp_all} \textit{Compare(A, B)}, we can find that the evaluations obtained in the many-shot regime are significantly better than those in the zero-shot regime.
					However, as mentioned before, the positional bias of GPT-4o may cause these results, so we performed a second comparison by swapping the positions. As shown in Figure~\ref{comp_all} \textit{Compare(B, A)}, we observe that as in-context examples increase, the evaluation results in the many-shot regime gradually turn around, that is, a higher winning rate. To obtain fairness results, we integrate the above two results (\textit{Compare(A, B)} and \textit{Compare(B, A)}), as shown in Figure~\ref{comp_final}. It can be seen that after mitigating the positional bias, the evaluation quality in the many-shot regime is still better than that in the zero-shot regime.
					
					\subsection{Revealing Symbol Bias}
					\citet{selection_bias} discover that these LLMs face selection bias, which means their prefer to select specific option IDs as answers (like “Option A”). 
					Meanwhile, \citet{song2024countingstars} find that when an LLM is stress-tested, it is easy to output some wrong information, which may be an increasing array related to the test data or an English alphabet sequence starting with "A". This phenomenon shows that LLMs favor answers with the symbol "A" rather than the symbol "B" (or the symbol "1" instead of the symbol "2").
					Therefore, an interesting question arises about: 
					\begin{itemize}
						\item \textit{{Do LLMs prefer to choose the answer with the symbol A or B?}}
					\end{itemize}
					
					To answer this question, we utilize the prompt template in Table~\ref{random_prompt} and replaced the symbols to be selected, conducting multiple experiments where LLMs chose symbols (set temperature=0.7 for all tested LLMs), as shown in Figure~\ref{random_pair}. It can be seen that the bias actually exists in LLMs when facing selection toward different symbols. Comparing the three LLMs at the same time, we find that the symbol bias of GPT-4o is relatively small, so in the pairwise comparison, we leverage GPT-4o as the evaluator. In addition, we find an interesting phenomenon that the position bias in the Claude3.5-sonnet and Gemini1.5-Pro are stronger than the symbol bias, and these LLMs tend to choose the options where the position is at the back. In contrast, GPT-4o tends to choose the symbols at the front. In other words, for Claude3.5-sonnet and Gemini1.5-Pro, the effect of position bias is heavier, and for GPT-4o the symbol bias impacts more.
					
					Inspired by our experiments, we swap the answers corresponding to symbols A and B, as shown in Figure~\ref{compare_prompt}. From Figure~\ref{comp_all} \textit{Compare(A$^\dagger$, B$^\dagger$)} and \textit{Compare(B$^\dagger$, A$^\dagger$)}, it can be seen that the evaluation results are different from the experiments. Actually, the results of \textit{Compare(A, B)} and \textit{Compare(A$^\dagger$, B$^\dagger$)} should be similar, and the results of \textit{Compare(B, A)} and \textit{Compare(B$^\dagger$, A$^\dagger$)} should be similar. This phenomenon shows that symbol bias does exist when adopting GPT-4o as an evaluator. 
					
\begin{figure}[t]
						\centering
						\includegraphics[scale=0.39]{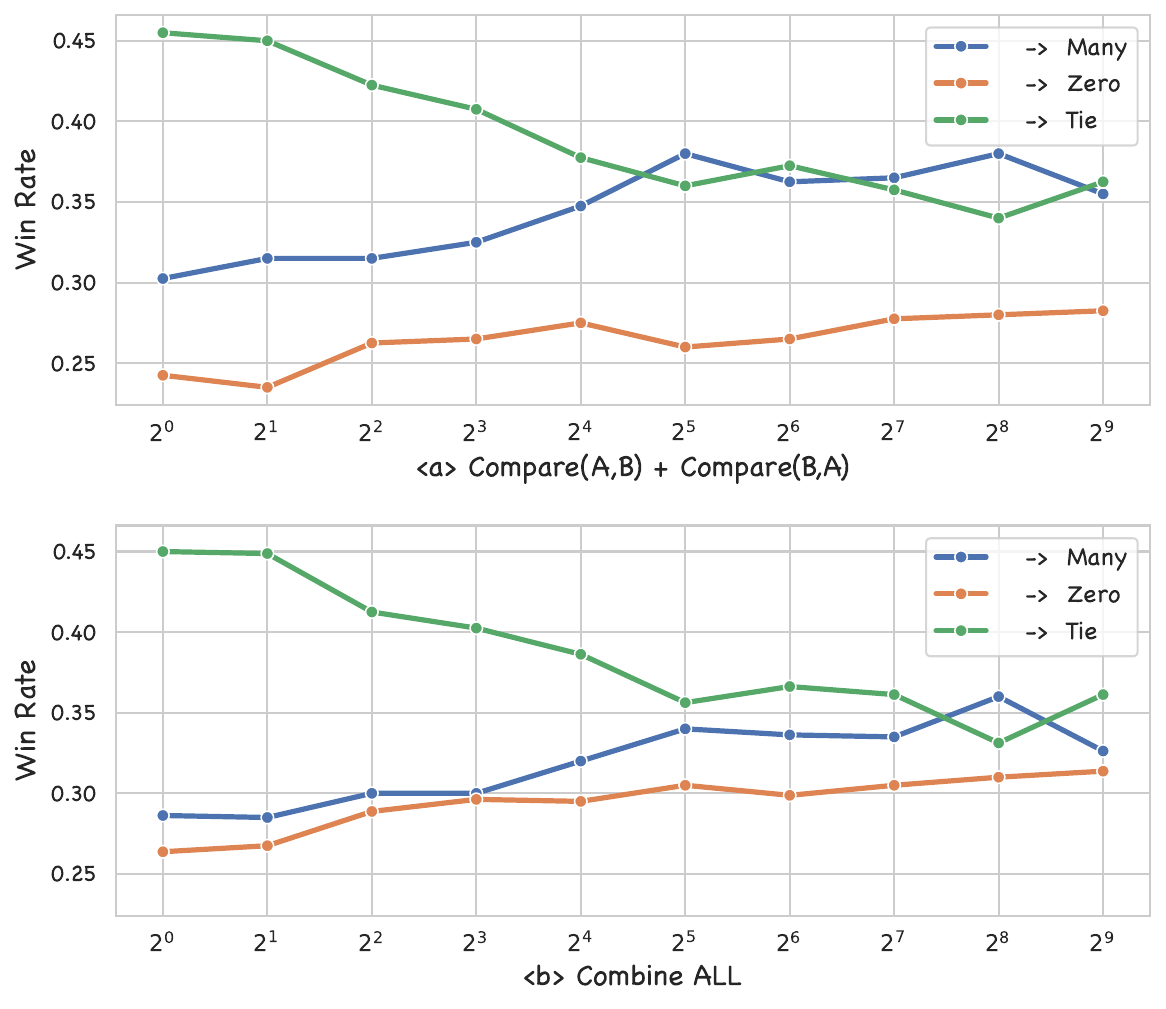}
						\caption{Comparison evaluation results after mitigating the biases via different approaches.}
						\label{comp_final}
					\end{figure}

					Recent research \cite{unfair} integrates the evaluation results of \textit{Compare(A, B)} and \textit{Compare(B, A)} to mitigate the positional bias, which motivates us to incorporate the evaluation results of \textit{Compare(A$^\dagger$, B$^\dagger$)} and \textit{Compare(B$^\dagger$, A$^\dagger$)} to reduce symbol bias. As presented in Figure~\ref{comp_final}, it can be seen that as in-context examples increase, the higher the win rate of the many-shot regime, which further verifies the effectiveness of the many-shot regime in helping GPT-4o as an evaluator.


					\section{Related Work}
					LLMs have exhibited remarkable general generation capabilities, positioning themselves as powerful assistants \cite{survey_llm, openai23}. With the rapid progression of LLMs, evaluating their proficiency in adhering to human instructions is imperative. Given the advanced capabilities of LLMs, researchers have begun adopting these models to evaluate the performance of LLMs in following human instructions \cite{Koo23, Liusie2023, liu2023, Zhu2023, lu2023, Fu2023, judge, unfair, chen}. Notably, the evaluation paradigm introduced by \citet{judge} has gained widespread adoption. However, LLMs as evaluators are revealed to have potential biases \cite{unfair, chen}, which leads to higher uncertainty and inconsistency during the evaluation using LLMs, questioning the validity of LLM evaluators.

					\section{Conclusion}
					In this work, we investigate and explore whether many-shot ICL helps LLMs as evaluators, such as GPT-4o. To this end, we designed two prompt templates, e.g., \textbf{M}any-\textbf{S}hot \textbf{w}ith \textbf{R}eference (\textbf{MSwR}) and \textbf{M}any-\textbf{S}hot with\textbf{o}ut \textbf{R}eference (\textbf{MSoR}). Experiments show that many-shot ICL can help the GPT-4o evaluator improve the consistency and quality of evaluation. Meanwhile, we also revealed symbol bias in LLMs when LLMs act as evaluators, and further proposed a simple yet effective approach to mitigate the symbol bias.


					\section{Limitations}
					Considering the trade-off between costs and benefits, we do not verify too many in-context examples in the experiments, such as thousands of examples. 
					Combining Figures~\ref{comp_long_gpt4} and Figure~\ref{comp_final}, it is not difficult to see that when the number of in-context examples increases to 256 and 512, although the consistency no longer improves, the evaluation quality has increased significantly.
					In addition, we consider that using GPT-4o as an evaluator in the many-shot regime is another evolution of the weak-to-strong strategy \cite{weak_strong}, which uses many zero-shot evaluation results to generate a better one.
					As the long-context capabilities of LLMs improve, adding more in-context examples may reveal more valuable phenomena for studying LLMs as evaluators in the future. 
					There are still many unexplored features of many-shot in-context learning. One possibility is that it can enable LLMs to simulate data constructed from shots, thereby increasing the diversity of synthetic data. Another possibility is to design benchmarks based on many-shot in-context examples and to assess the long-context LLMs based on the capabilities of understanding and imitating many shots.

					\section*{Acknowledgments}
					We thank the three anonymous reviewers for carefully reading our paper and their insightful comments and suggestions.
					
					\bibliography{custom}

				\end{document}